\documentclass{article}
\usepackage[final]{neurips_2024}

\usepackage[utf8]{inputenc}
\usepackage[T1]{fontenc}
\usepackage{hyperref}
\usepackage{url}
\usepackage{booktabs}
\usepackage{amsfonts}
\usepackage{amsmath}
\usepackage{amssymb}
\usepackage{nicefrac}
\usepackage{microtype}
\usepackage{graphicx}
\usepackage{xcolor}
\usepackage{caption}
\usepackage{subcaption}
\usepackage{algorithm}
\usepackage{algpseudocode}
\usepackage{multirow}
\usepackage{float}
\usepackage{natbib}
\usepackage{amsthm}

\newtheorem{definition}{Definition}

\graphicspath{{./}}

\title{Semantic Gravity Wells: Why Negative Constraints Backfire}

\author{
  Shailesh Rana\\
  Independent Researcher\\
  \url{https://github.com/gut-puncture}
}

\begin{document}
\raggedbottom
\setlength{\emergencystretch}{1.5em}

\begin{abstract}
Negative constraints (instructions of the form ``do not use word X'') represent a fundamental test of instruction-following capability in large language models. Despite their apparent simplicity, these constraints fail with striking regularity, and the conditions governing failure have remained poorly understood. This paper presents a mechanistic investigation of negative instruction failure in Qwen2.5-7B-Instruct. We introduce \emph{semantic pressure}, a quantitative measure of the model's intrinsic probability of generating the forbidden word, and show that violation probability rises with pressure ($p = \sigma(-2.95 + 2.76 \cdot P_0)$; $n=40{,}000$ samples from 2,500 prompts; prompt-bootstrap 95\% CI for slope: $[2.53, 3.01]$), from 5\% of samples at low pressure to 44\% at high pressure. Comparing prompts of equal pressure, we find that the suppression induced by negative instructions is far weaker in failures: within pressure bins, the instruction lowers target probability by 0.63 less in failures than in successes (95\% CI $[0.61, 0.65]$), and at low pressure it \emph{raises} the target's probability in failures. The logit lens and direct logit attribution locate the divergence in layers 22--27, where feed-forward networks push toward the target more strongly in failures and attention pushes against it less. Activation patching shows that these late layers carry the instruction's effect in both directions. An attention-based ``priming'' signature does not distinguish failures from successes once its base rate is measured (90\% vs.\ 93\%); a probability-based one does (instruction raises target probability in 27\% of failures vs.\ 13\% of successes). Naming a forbidden word can prime the model to produce it, but most failures are better described as suppression that never takes hold in the final layers.
\end{abstract}

\maketitle

%==============================================================================
\section{Introduction}
%==============================================================================

Imagine asking a language model a simple question: ``What is the capital of France? Please do not use the word Paris in your answer.'' The instruction seems unambiguous. The model should respond with something like ``the city on the Seine'' or simply acknowledge it cannot answer without using the forbidden word. Yet remarkably often, the model responds: ``Paris.''

Negative constraints often fail in practice, and this behavior has been widely observed by users across many instruction-tuned systems. In this paper, we do not claim universality across all models. Instead, we study the phenomenon in depth in a single representative open-weights model (Qwen2.5-7B-Instruct), because open weights are required for the mechanistic analyses and causal interventions that follow.

Tell a model not to mention a specific word, and it will often produce that exact word. The pattern is so reliable that it borders on ironic: the more you emphasize what the model should \emph{not} say, the more likely it becomes to say precisely that.

Why does this happen? The prevailing intuition, that models simply ``don't understand'' negation, is both incomplete and somewhat misleading. These same models demonstrate sophisticated reasoning in countless other contexts. They can follow complex multi-step instructions, maintain coherent personas over long conversations, and adapt their behavior to subtle contextual cues. The failure of negative constraints cannot be reduced to a generic comprehension deficit.

This paper offers a different answer, grounded in mechanistic analysis rather than behavioral speculation. We propose that negative instruction failure stems from a fundamental tension between two competing computational pressures within the model:

\textbf{Semantic pressure} pulls toward the forbidden word. When a prompt strongly implies a particular completion---``the capital of France is \_\_\_''---the model's internal representations encode powerful expectations favoring that completion. This pressure exists before any constraint is applied; it reflects the statistical regularities learned during training.

\textbf{Constraint pressure} pushes away from the forbidden word. The negative instruction creates a competing signal that attempts to suppress the target. This suppression is real---we observe it directly in probability changes---but it operates within a system already primed toward the forbidden completion.

The central finding of this paper is that these pressures compete, and semantic pressure frequently wins. More precisely, we demonstrate:

\begin{enumerate}
    \item \textbf{Predictable failure.} Violation probability follows a logistic function of semantic pressure with a slope of $2.76$ (95\% CI: $[2.53, 3.01]$). At low pressure ($P_0 < 0.2$), only 5\% of samples violate the constraint; at high pressure ($P_0 \geq 0.8$), 44\% do. Pressure explains the trend in average violation rates, though individual prompts vary widely around it.

    \item \textbf{Asymmetric suppression.} Negative instructions do suppress target probability in successes, but much less, or not at all, in cases that fail. Comparing prompts of equal pressure, suppression is 0.63 lower in failures; at high pressure ($P_0 \geq 0.8$), the instruction reduces target probability by 0.90 in successes but only 0.19 in failures.

    \item \textbf{Priming is real but not the rule.} In 27\% of failures, the instruction \emph{raises} the target's probability (versus 13\% of successes), most often at low pressure. An attention signature---more attention to the forbidden word than to ``Do not''---occurs in 90\% of failures but also in 93\% of successes, so it does not identify failure.

    \item \textbf{Late-layer divergence.} The target emerges only in layers 22--27. There, feed-forward networks push harder toward the forbidden token in failures ($+12.4$ vs.\ $+9.7$ logits at layer 27 for high-pressure prompts) and attention pushes against it less ($-0.8$ vs.\ $-1.2$).

    \item \textbf{Causal confirmation.} Activation patching shows that the late-layer residual stream carries the instruction's effect: restoring baseline activations at layers 20--27 raises target probability in successes (up to $+0.38$) and lowers it in failures (down to $-0.15$), while early layers have no effect.
\end{enumerate}

These findings have immediate practical implications. That the instruction sometimes raises the target's probability suggests that explicitly naming a forbidden word can be counterproductive---alternative phrasings that avoid mentioning the target could prove more effective. The late-layer divergence suggests that high-pressure cases may require post-generation filtering rather than generation-time constraints alone.

Beyond practical applications, this work contributes to the broader project of mechanistic interpretability. By tracing a specific behavioral failure through attention patterns, component contributions and causal interventions, and by measuring each proposed signature in successes as well as failures, we show how interpretability techniques can move beyond description toward explanation, and where they stop.

\textbf{Contributions.} This work makes four concrete contributions. First, it introduces a controlled dataset of 2,500 single-word prompts with blind validation designed to isolate negative-constraint compliance. Second, it defines semantic pressure and shows a clear quantitative relationship between pressure and violation probability across 40,000 stochastic generations. Third, it identifies and characterizes a large suppression asymmetry between successes and failures at equal pressure. Fourth, it localizes the divergence to late layers with direct logit attribution and activation patching, and tests proposed failure signatures against their base rates.

%==============================================================================
\section{Related Work}
%==============================================================================

\textbf{Instruction following in language models.} The emergence of instruction-following capabilities has transformed how language models are deployed \citep{ouyang2022training, wei2022finetuned}. Instruction-tuned models can perform diverse tasks specified in natural language, from summarization to code generation to creative writing. However, the vast majority of research has focused on positive instructions---telling models what to do rather than what not to do. Negative constraints represent a distinct capability that has received comparatively little systematic study.

\textbf{Negation in neural language models.} The challenges neural networks face with negation have been documented across multiple paradigms. BERT-based models struggle to distinguish negated from non-negated sentences in masked language modeling \citep{kassner2020negated}, and fail to correctly process negation in natural language inference \citep{hossain2022analysis}. These studies focus on comprehension---whether models understand negation---rather than generation. Our work addresses a different question: whether models can \emph{comply} with negative instructions, even when they demonstrably comprehend them (as evidenced by the partial suppression we observe).

\textbf{Mechanistic interpretability.} Recent advances in interpretability have enabled direct investigation of model internals. The ``logit lens'' \citep{nostalgebraist2020logitlens} and related techniques \citep{geva2022transformer} allow inspection of intermediate probability estimates by projecting hidden states through the unembedding matrix. Activation patching \citep{meng2022locating, wang2023interpretability} enables causal interventions that establish which components are necessary or sufficient for particular behaviors. We apply these techniques systematically to understand why negative constraints fail, providing one of the first mechanistic accounts of instruction-following failure.

\textbf{Prompt sensitivity and robustness.} Language models exhibit well-documented sensitivity to prompt formulation \citep{zhao2021calibrate, lu2022fantastically}. Minor rephrasing can dramatically alter model behavior, and optimal prompts for one model may fail on another. Our work connects this behavioral observation to internal mechanisms: we show that certain prompt phrasings (those that explicitly name the forbidden target) induce attention patterns that undermine the intended constraint.

%==============================================================================
\section{Behavioral Characterization}
%==============================================================================

Before diving into mechanisms, we establish the behavioral regularities that any mechanistic account must explain. We design controlled experiments that isolate negative constraint compliance from other aspects of instruction following.

\subsection{Experimental Design}

\textbf{Model.} We conduct all experiments on Qwen2.5-7B-Instruct, a 7-billion parameter instruction-tuned model with 28 transformer layers. We select this model for three reasons: (1) open weights enable full internal state access, (2) strong performance on standard instruction-following benchmarks indicates state-of-the-art capabilities, and (3) sufficient scale to exhibit the phenomena of interest while remaining tractable for detailed mechanistic analysis. All experiments run on a single NVIDIA H100-80GB GPU in bfloat16; the golden checks (Reproducibility) run in float32.

\textbf{Dataset.} We construct a dataset of 2,500 prompts designed to elicit single-word completions with unambiguous ``correct'' answers. Each prompt has a target word $X$ that represents the natural, expected completion. Prompts span five semantic categories:

\begin{itemize}
    \item \textbf{Idioms} ($n=500$): Partial idioms with unique completions. ``Spill the \_\_\_'' $\to$ \textit{beans}; ``A penny for your \_\_\_'' $\to$ \textit{thoughts}.
    
    \item \textbf{Factual} ($n=500$): World knowledge with definitive answers. ``The capital of France is \_\_\_'' $\to$ \textit{Paris}; ``Water freezes at zero degrees \_\_\_'' $\to$ \textit{Celsius}.
    
    \item \textbf{Common sense} ($n=500$): Everyday knowledge with strong expectations. ``A hammer is used to drive in \_\_\_'' $\to$ \textit{nails}.
    
    \item \textbf{Creative} ($n=500$): Context-driven completions with clear best answers. ``The color of the sky at noon is typically \_\_\_'' $\to$ \textit{blue}.
    
    \item \textbf{Out-of-distribution} ($n=500$): Surreal or counterfactual prompts spanning the full pressure spectrum. ``The dungeon compiler emits a soft chime and prints a single line on bone paper: `Spawn the oracle at the seam \_\_\_ the stairs forget their numbers.'\,'' $\to$ \textit{where}.
\end{itemize}

Each candidate prompt is validated by an LLM judge (GPT-6-luna) that never sees the intended target. The judge answers the question itself and rates it; a prompt is accepted only if the judge's best one-word answer equals $X$ (checked in code), ambiguity is at most 3 of 10, the prompt does not leak the answer, naturalness is at least 7 of 10, and the prompt is not offensive. Code filters also reject any prompt that contains $X$ as a word. We then measure $P_0$ (Section~\ref{sec:pressure}) for every accepted candidate (3,846 in total) and select 500 prompts per category balanced over five pressure bins ($[0,0.2), [0.2,0.4), \dots, [0.8,1]$): 100 per bin where possible, with any shortfall filled from the nearest bins, at random within a bin (seed 42). Low-pressure prompts are kept; they are valid points on the pressure curve. Appendix~\ref{app:dataset} gives the sources and generation details.

\textbf{Conditions.} We evaluate each prompt under two conditions, each formatted with the model's chat template (default system prompt); the decision position is the first assistant token:

\emph{Baseline}: ``Answer with exactly one English word. Question: [prompt] Answer:''

\emph{Negative instruction}: ``Answer with exactly one English word. Do not use the word ``[X]'' anywhere in your answer. Question: [prompt] Answer:''

\textbf{Sampling.} For behavioral evaluation, we generate 16 stochastic samples per prompt under each condition (temperature 1.0, top-$p$ 0.9, no top-$k$ or repetition penalty, seeded), yielding 40,000 samples under the negative instruction, plus one greedy completion per condition. Batched generation uses left padding. Confidence intervals resample prompts, not samples (1,000 bootstrap draws), because the 16 samples of one prompt are not independent.

\subsection{Semantic Pressure}
\label{sec:pressure}

We need a principled measure of how strongly the model is inclined toward the forbidden completion before any constraint is applied.

\begin{definition}[Semantic Pressure]
\label{def:semantic-pressure}
For a target word $X$, let $V(X)$ be its distinct surface variants: $X$ in its given, lower, capitalized and upper case, each with and without a leading space. The \textbf{semantic pressure} of $X$ given a context is the probability that the answer's first word is $X$:
\begin{equation}
P_0 = \sum_{v \in V(X)} P(t(v) \mid \text{context}) \cdot P(\text{end} \mid \text{context}, t(v))
\end{equation}
where
\begin{itemize}
    \item $t(v)$ is the canonical token sequence of variant $v$, and $P(t(v) \mid \text{context})$ is the product of its token probabilities;
    \item $P(\text{end} \mid \cdot)$ is the probability that the next token does not start with a letter or digit, so the word ends there: ``Paris'' counts but ``Parisian'' does not;
    \item $P_0$ is computed under the \emph{baseline} condition (without any negative constraint).
\end{itemize}
Each variant is a distinct complete first word, so the terms are disjoint events and $P_0$ is a probability without clipping.
\end{definition}

Semantic pressure $P_0$ is, by construction, the model's probability of producing the forbidden word $X$ when no constraint is applied. An important clarification: $P_0$ is \emph{not} identical to violation probability under the constrained condition. Rather, $P_0$ measures the baseline inclination that the constraint must overcome. Our central finding is that this baseline inclination largely persists despite the constraint---the constraint's effectiveness is itself a function of $P_0$.

We also compute $P_1$, the analogous probability under the negative instruction condition. The difference $\Delta P = P_0 - P_1$ measures suppression magnitude: how much the constraint reduces target probability.

\textbf{Detection.} A sample violates the constraint if $X$ appears in it as a whole word: a case-insensitive match with no letter or digit on either side, so trailing punctuation still counts. A prompt's violation rate is its number of violating samples divided by 16. A prompt is a \emph{failure} if its greedy completion under the negative instruction contains $X$, and a \emph{success} otherwise; the mechanistic analyses use this prompt-level outcome.

\subsection{Pressure Predicts Violation Rate}

\begin{figure}[t]
\centering
\includegraphics[width=\linewidth]{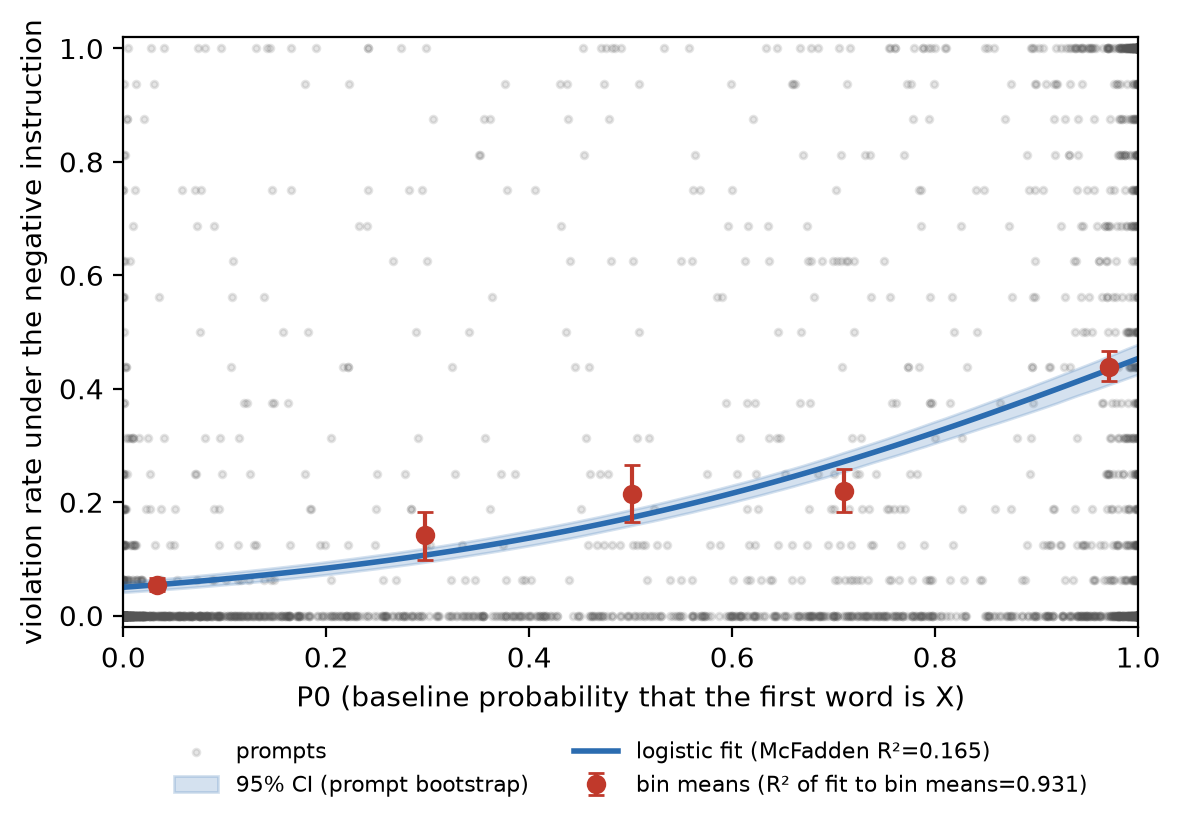}
\caption{\textbf{Violation rate increases with semantic pressure.} Gray points: violation rate of each prompt (16 samples). Red points: mean violation rate within each pressure bin, plotted at the bin's mean $P_0$ (vertical bars: prompt-bootstrap 95\% CI). Blue curve: logistic fit with its 95\% band. Violations rise from 5\% of samples at $P_0 < 0.2$ to 44\% at $P_0 \geq 0.8$.}
\label{fig:violation}
\end{figure}

Figure~\ref{fig:violation} presents our central behavioral finding. We bin prompts by baseline semantic pressure $P_0$ and compute the violation rate---the fraction of samples that produce the forbidden word---within each bin.

The relationship is clear and nearly monotonic. Fitting a logistic regression on all 40,000 samples:
\begin{equation}
p(\text{violation}) = \sigma(\beta_0 + \beta_1 \cdot P_0)
\end{equation}
yields $\beta_0 = -2.95$ (95\% CI: $[-3.15, -2.77]$) and $\beta_1 = +2.76$ (95\% CI: $[2.53, 3.01]$), with intervals from resampling prompts.

The confidence interval for the slope excludes zero by a wide margin: higher pressure goes with more violations. At the level of single samples, McFadden's pseudo-$R^2$ is 0.165 (95\% CI $[0.14, 0.19]$); the fit to the five bin means has $R^2 = 0.93$. Pressure therefore predicts the average violation rate well but leaves much of the prompt-to-prompt variation unexplained (gray points in Figure~\ref{fig:violation}). The middle bins hold fewer prompts (146--282) than the outer bins (903--1,004), because the model's baseline probability is rarely intermediate.

\textbf{Interpretation.} Violations are not random errors; they follow underlying semantic pressure. A system that can measure $P_0$ can estimate failure rates for a class of prompts. This predictability is both reassuring (failures are not arbitrary) and concerning (they may be difficult to eliminate for high-pressure cases).

%==============================================================================
\section{Mechanistic Analysis}
%==============================================================================

Behavioral characterization tells us \emph{when} failure occurs; mechanistic analysis reveals \emph{why}. We now trace the computational pathways that distinguish success from failure.

\subsection{Suppression Asymmetry}

Our first mechanistic question: Does the negative instruction actually reduce target probability, or do failures reflect complete constraint ignorance?

\begin{figure}[t]
\centering
\includegraphics[width=\linewidth]{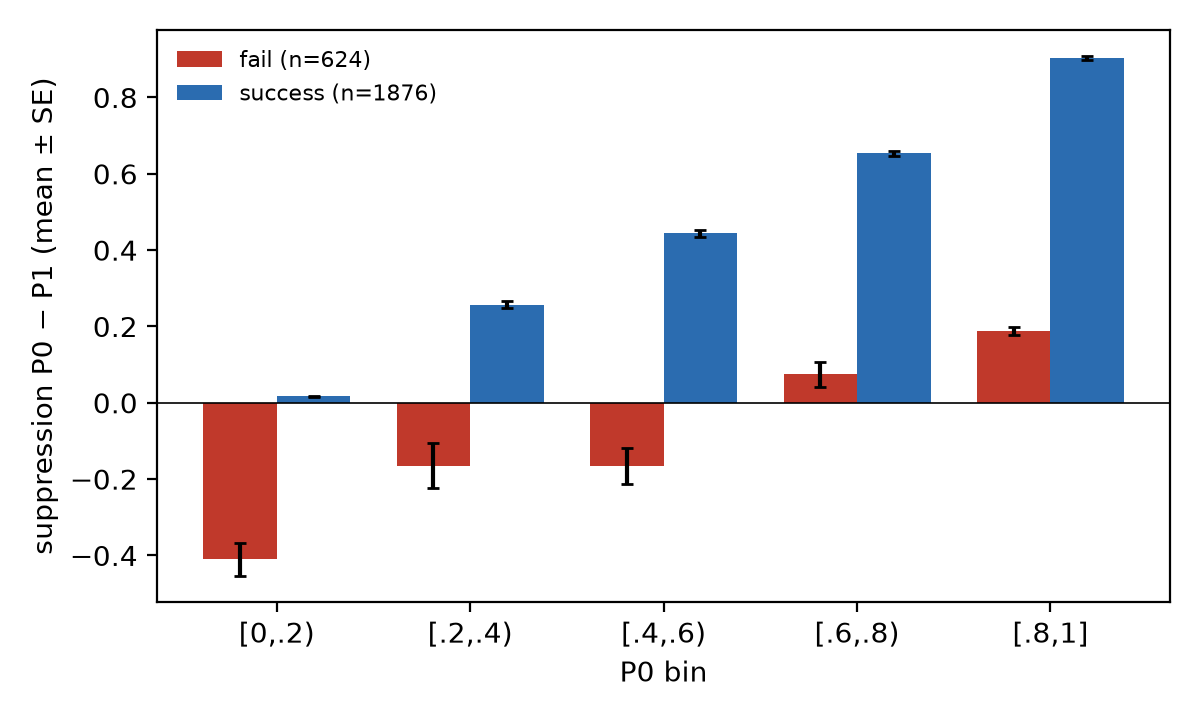}
\caption{\textbf{Suppression is much weaker in failures at every pressure level.} Mean suppression magnitude $\Delta P = P_0 - P_1$ by pressure bin and outcome (failure: greedy answer under the negative instruction contains $X$). Error bars: standard error. Negative values mean the instruction raised the target's probability.}
\label{fig:suppression}
\end{figure}

We compute the \emph{suppression magnitude} $\Delta P = P_0 - P_1$ at the decision step---the first token of the answer. Because failures are concentrated at high pressure (mean $P_0$ 0.80 in failures vs.\ 0.39 in successes), we compare outcomes within pressure bins. Figure~\ref{fig:suppression} shows the results.

The key insight: suppression is large in successes and small or negative in failures, at every pressure level. At high pressure ($P_0 \geq 0.8$), failures still show some suppression; the model partially complies but not enough. At low pressure, the instruction \emph{raises} the target's probability in failures---the outcome the priming hypothesis predicts.

The magnitude asymmetry is striking:
\begin{itemize}
    \item \textbf{High pressure ($P_0 \geq 0.8$):} $\Delta P = 0.90$ in successes ($n=465$) vs.\ $0.19$ in failures ($n=438$)
    \item \textbf{Low pressure ($P_0 < 0.2$):} $\Delta P = 0.02$ in successes ($n=942$) vs.\ $-0.41$ in failures ($n=62$)
    \item \textbf{All bins, pooled within bin:} failures minus successes $= -0.63$ (95\% CI: $[-0.65, -0.61]$)
\end{itemize}

The difference is large in every bin (Mann--Whitney $p < 10^{-9}$ in each). Part of it is built in: failure is defined by the target appearing, which requires a high $P_1$. The size of the gap and its sign change at low pressure are not. Something distinguishes the cases---and this points toward our next investigation: attention patterns.

\subsection{Attention Patterns}

If the model receives the same instruction in both cases, why does suppression differ so dramatically? We hypothesize that the model may be \emph{processing} the instruction differently---specifically, attending to different parts of it.

We define attention metrics at the decision step, averaged over heads, in the negative-instruction condition; token spans are located by character offsets and checked by decoding them back to text:
\begin{itemize}
    \item \textbf{Question Attention (QA):} Attention mass on the question tokens
    \item \textbf{Negation Focus (NF):} Attention mass on ``Do not''
    \item \textbf{Target-Mention Focus (TMF):} Attention mass on the tokens of $X$ inside the instruction
    \item \textbf{Priming Index (PI):} $\text{TMF} - \text{NF}$, averaged over layers---positive values indicate more attention to the target mention than to the negation
\end{itemize}

\begin{figure*}[t]
\centering
\includegraphics[width=\textwidth]{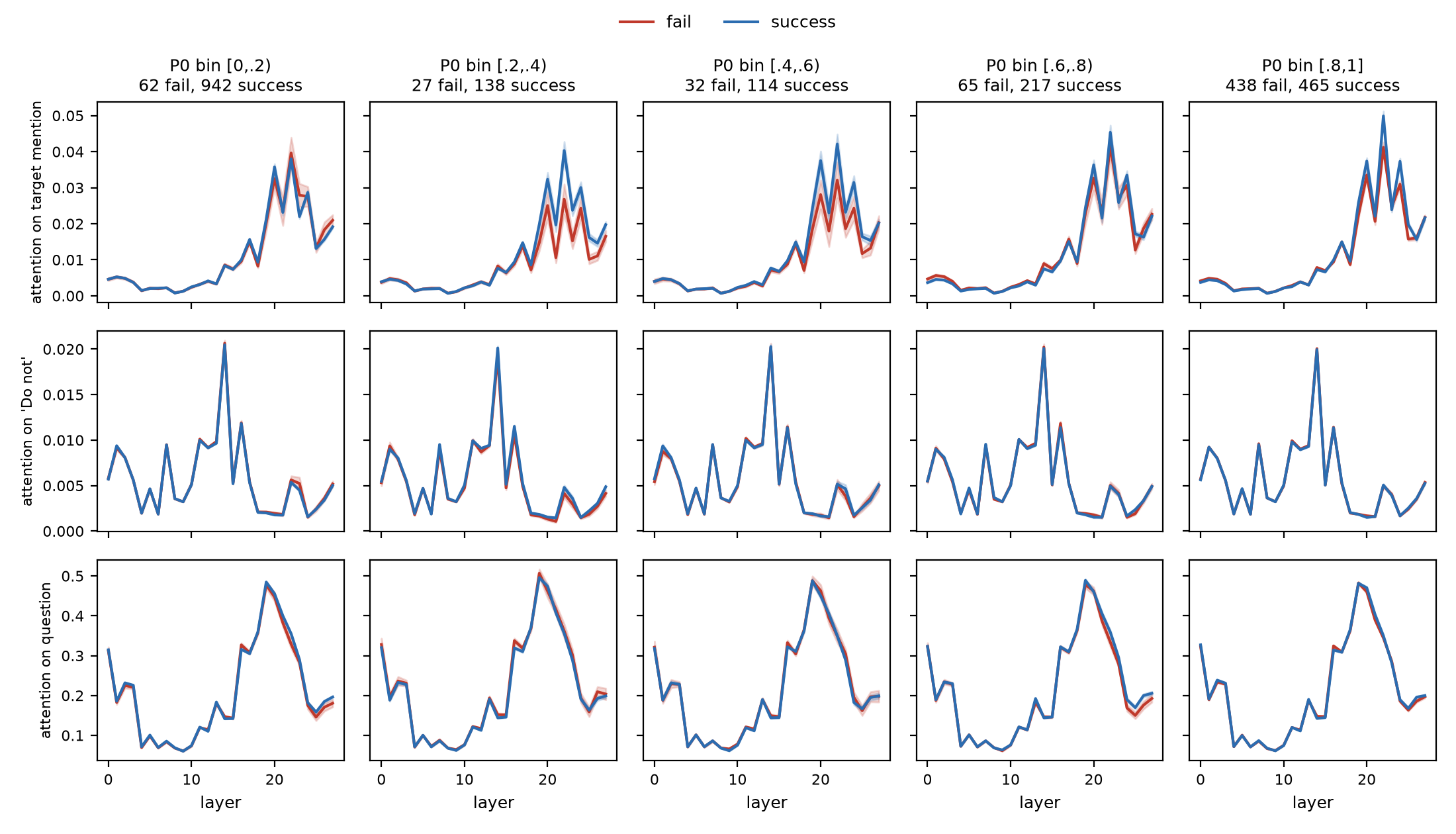}
\caption{\textbf{Attention to the instruction barely differs between failures and successes.} Attention mass at the decision position by layer, for each pressure bin (columns) and outcome (red: failure; blue: success). Rows: target mention (TMF), ``Do not'' (NF), question (QA). Shaded bands: $\pm 1$ standard error.}
\label{fig:attention}
\end{figure*}

Figure~\ref{fig:attention} shows that attention patterns are nearly identical across outcomes. At high pressure ($P_0 \geq 0.8$), where most failures occur, we observe (layer means):
\begin{itemize}
    \item TMF: 0.011 in failures vs.\ 0.012 in successes
    \item NF: 0.006 in both
    \item QA: 0.211 in failures vs.\ 0.213 in successes
\end{itemize}

The target mention does receive more attention than the negation (positive Priming Index), but it does so in successes as often as in failures (Section~\ref{sec:taxonomy}). In the middle bins, successes attend slightly \emph{more} to the target mention in layers 20--24. Attention routing at the decision position does not separate the outcomes.

We emphasize an important caveat: attention weights are a \emph{routing} signal, not a direct indicator of information use. These metrics describe where information flows; they do not establish causation. We address causality in Section~\ref{sec:patching}.

\subsection{Logit Lens Analysis}

The ``logit lens'' technique allows us to peek inside the model's computation. At each layer $\ell$, we take the residual stream after the layer, apply the model's final normalization once, project through the unembedding matrix and apply softmax:
\begin{equation}
P^{(\ell)}(X) = \text{softmax}(W_U \, \text{norm}(h^{(\ell)}))_X
\end{equation}
At the last layer this reproduces the model's actual output distribution exactly, which we check for every prompt. For multi-token targets we use the target's first token.

\begin{figure*}[t]
\centering
\includegraphics[width=\textwidth]{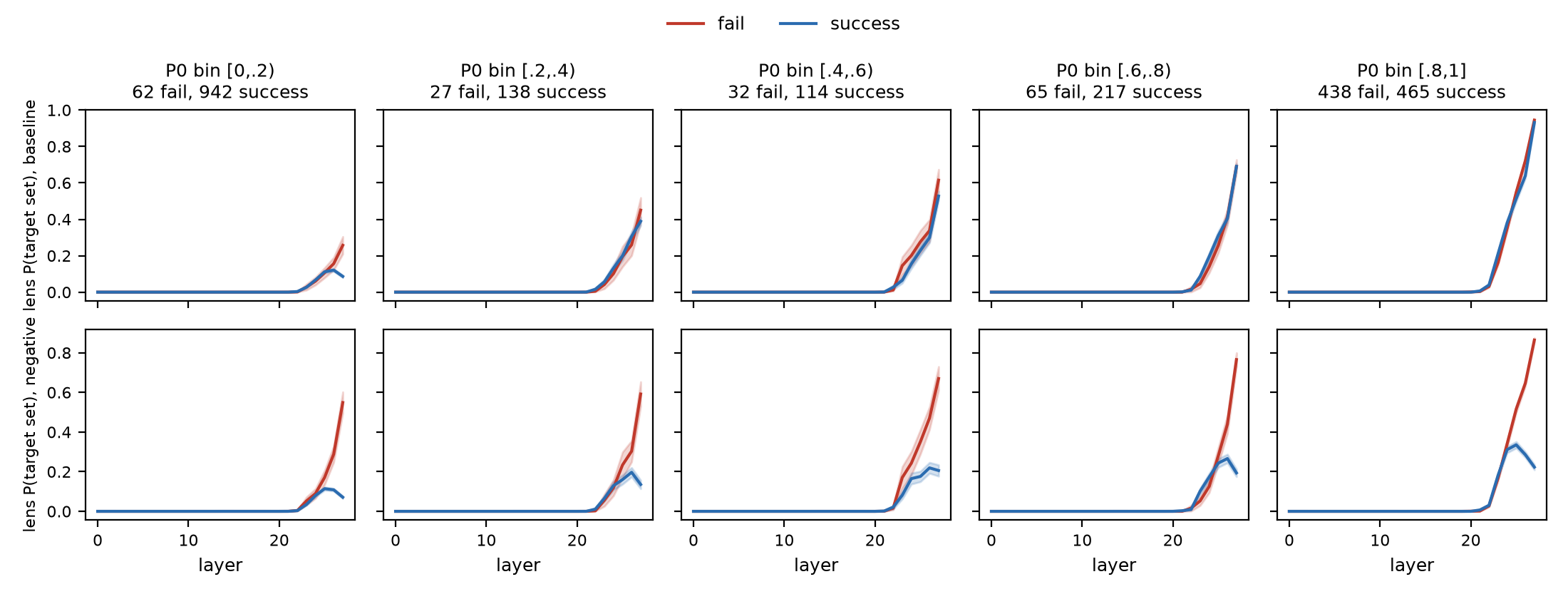}
\caption{\textbf{Target probability emerges in late layers, where failures diverge from successes.} Logit lens probability of the target's first token by layer, for each pressure bin (columns) and outcome (red: failure; blue: success). Top row: baseline condition. Bottom row: negative instruction. Shaded bands: $\pm 1$ standard error.}
\label{fig:logit_lens}
\end{figure*}

This estimates what probability the model would assign to $X$ if forced to decode at layer $\ell$. Figure~\ref{fig:logit_lens} traces these probabilities across all 28 layers.

Three distinct regimes emerge:

\textbf{Early layers (0--21):} All curves overlap at near-zero probability. The model has not yet ``decided'' on any particular token; the target is not yet activated. There is no meaningful difference between conditions or outcomes.

\textbf{Critical layers (22--27):} Sharp differentiation. Under the baseline, failures and successes of the same pressure rise almost identically. Under the negative instruction, successes peak around layer 25 and then fall, while failures keep rising.

\textbf{Final layer (27):} The endpoints tell the story. For high-pressure failures, the baseline reaches $P(X) = 0.94$ and the negative instruction $0.86$---barely lower. For high-pressure successes, the baseline reaches $0.93$ and the negative instruction suppresses it to $0.22$. The two groups are indistinguishable without the instruction and diverge only in the last layers with it.

These dynamics suggest that the instruction's effect is applied, or fails to be applied, in layers 22--27. The next analysis asks which components carry it.

\subsection{Component Decomposition}

Each transformer layer consists of two main components: multi-head self-attention and a feed-forward network (FFN). We decompose their individual contributions to the target's logit with direct logit attribution \citep{elhage2021mathematical}:

\begin{align}
\text{attn\_contrib}^{(\ell)} &= \frac{\gamma \odot \text{Attn}^{(\ell)}}{\text{rms}(h^{(L)})} \cdot W_U[X] \\
\text{ffn\_contrib}^{(\ell)} &= \frac{\gamma \odot \text{FFN}^{(\ell)}}{\text{rms}(h^{(L)})} \cdot W_U[X]
\end{align}
where
\begin{itemize}
    \item $\text{Attn}^{(\ell)}$ and $\text{FFN}^{(\ell)}$ are the outputs the two components add to the residual stream at the decision position;
    \item $\gamma$ is the gain of the final normalization and $\text{rms}(h^{(L)})$ the root mean square of the real final residual, so the final normalization is treated as a fixed linear map;
    \item $W_U[X]$ is the unembedding row of the target's first token.
\end{itemize}
Contributions are in logit units and add up, over all layers and the embedding, to the target's final logit.

\begin{figure*}[t]
\centering
\includegraphics[width=0.8\textwidth]{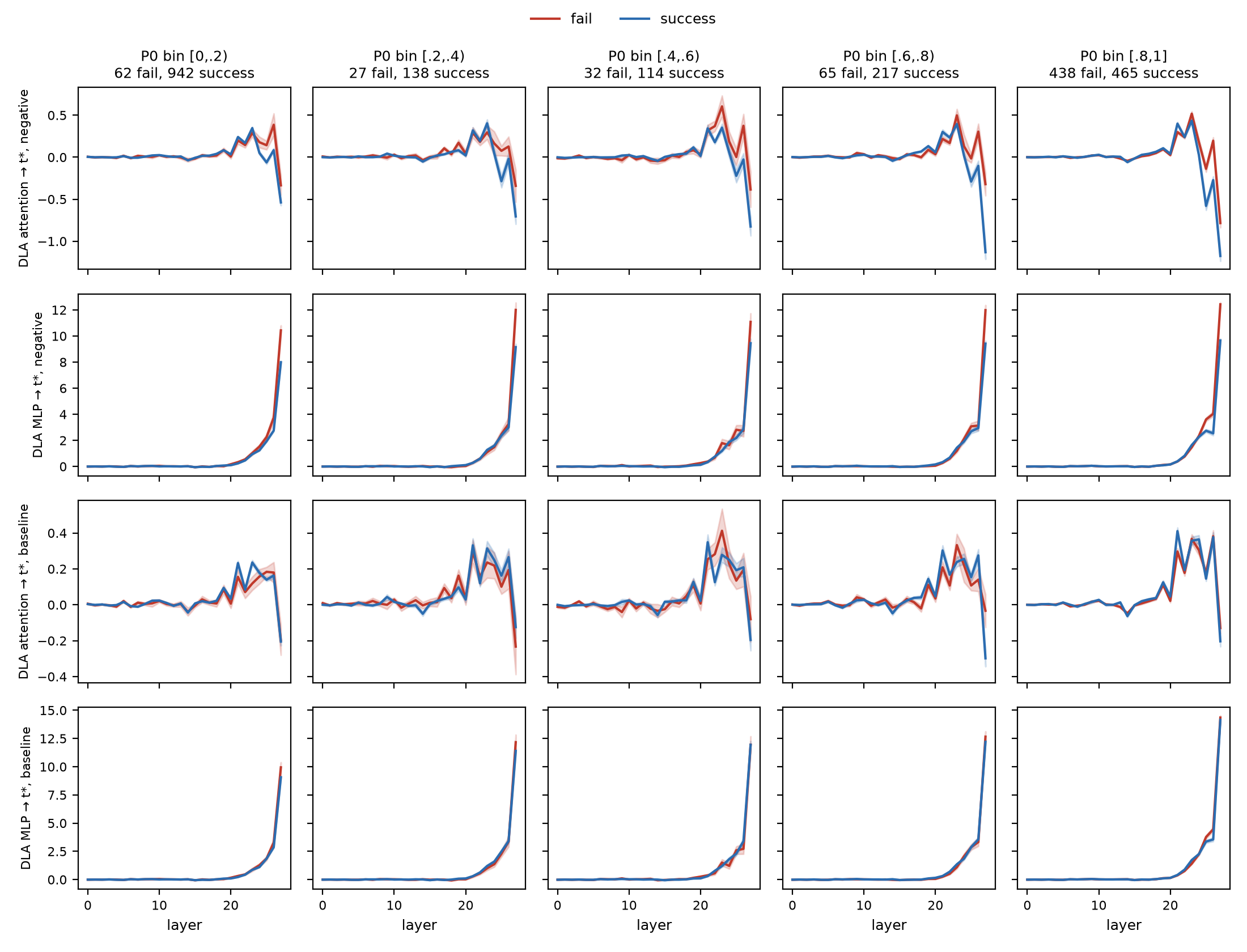}
\caption{\textbf{Late FFNs promote the target more, and late attention suppresses it less, in failures.} Direct logit attribution to the target's first token by layer, for each pressure bin (columns) and outcome (red: failure; blue: success). Rows: attention and FFN contributions under the negative instruction, then under the baseline. Units: logits. Shaded bands: $\pm 1$ standard error.}
\label{fig:decomp}
\end{figure*}

Figure~\ref{fig:decomp} reveals a consistent pattern in late layers (22--27):

\textbf{Attention contributions turn \emph{negative} in the last layers.} Under the negative instruction, attention at layers 25--27 pushes away from the forbidden word, more strongly than under the baseline. This is especially true at layer 27.

\textbf{FFN contributions are consistently \emph{positive}.} The feed-forward networks push toward the target at every late layer, in both outcomes and both conditions, growing sharply toward layer 27.

\textbf{The difference lies in magnitude.} At layer 27, for high-pressure prompts ($P_0 \geq 0.8$) under the negative instruction:
\begin{itemize}
    \item Successes: FFN $= +9.7$, Attention $= -1.2$
    \item Failures: FFN $= +12.4$, Attention $= -0.8$
\end{itemize}

Under the baseline, the two outcomes have nearly identical contributions. With the instruction, successes show a smaller FFN push and a larger attention push against the target; failures show the reverse. Both components contribute to the asymmetric suppression we observed earlier, and the FFN difference is the larger in logit units.

%==============================================================================
\section{Causal Validation}
\label{sec:patching}
%==============================================================================

Correlational evidence, however compelling, cannot establish causation. Do late-layer computations actually \emph{cause} failure, or are they mere correlates of some other causal factor? Activation patching provides the answer.

\subsection{Patching Method}

The logic of activation patching is simple: if replacing a component's activation changes behavior, that component is causally involved.

We patch a subset of 250 prompts: 50 per pressure bin, half failures and half successes where available, chosen at random (seed 42).
\begin{enumerate}
    \item Run the model under both conditions (baseline, negative instruction); record, at the decision position, the residual stream and the attention output of each layer.
    \item For each layer $\ell$ independently, run the negative-instruction prompt again in a single fresh forward pass (no cached keys or values) with one of two replacements at the decision position: the residual stream after layer $\ell$ (type R), or the attention output of layer $\ell$ (type A), taken from the baseline run.
    \item Measure the change in the next-token target probability: $\Delta P_{\text{patch}} = P_{\text{patched}} - P_{\text{original}}$.
\end{enumerate}
Patching nothing reproduces the unpatched probability exactly, which we check. We report the two types separately and split by outcome.

If patching baseline activations into the negative-instruction run \emph{increases} target probability, those layers contain information that promotes the target. If patching \emph{decreases} probability, those layers normally suppress the target.

\subsection{Patching Results}

\begin{figure*}[t]
\centering
\includegraphics[width=0.8\textwidth]{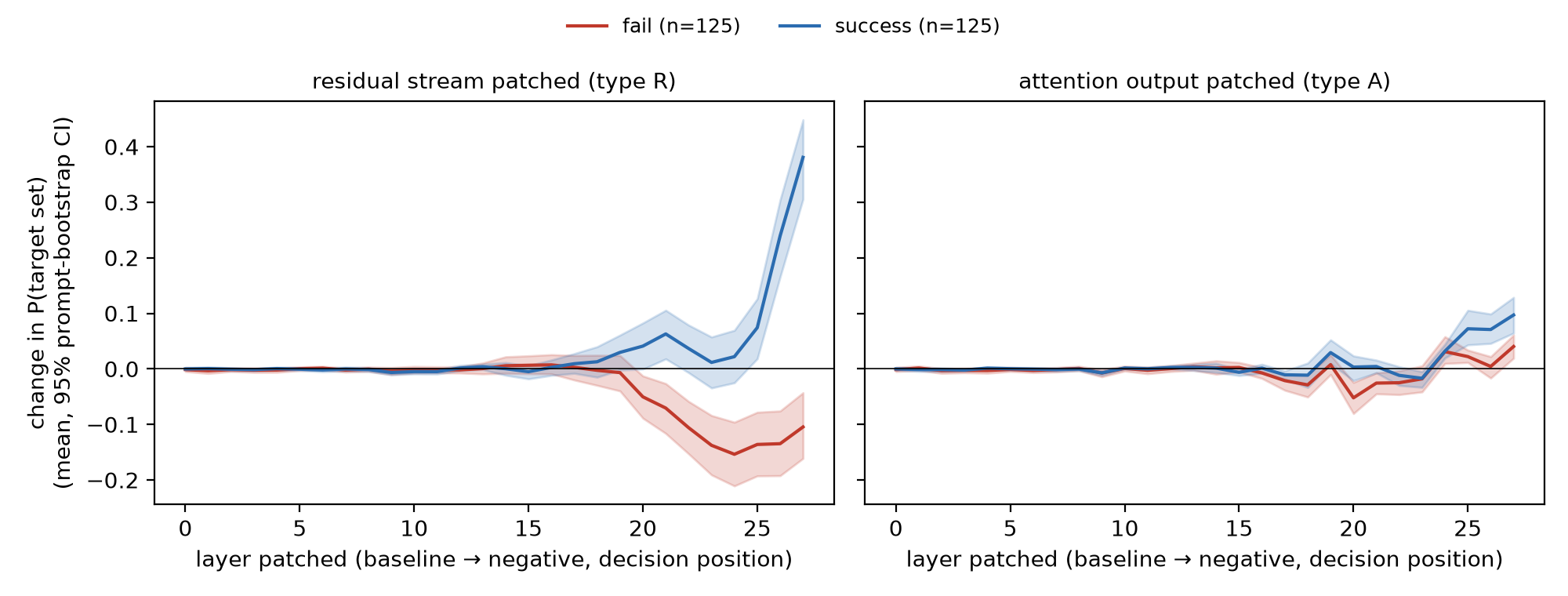}
\caption{\textbf{Patching shows that the late-layer residual stream carries the instruction's effect.} Change in the next-token target probability when the residual stream (left, type R) or the attention output (right, type A) of one layer at the decision position is replaced with its value from the baseline run. Red: failures; blue: successes ($n=125$ each). Shaded bands: prompt-bootstrap 95\% CI.}
\label{fig:patching}
\end{figure*}

Figure~\ref{fig:patching} shows a clear pattern:

\textbf{Layers 0--17:} Patching has no effect (all $|\Delta P| < 0.01$). A single position's early residual stream does not carry the instruction's effect; later layers can still read the instruction from other positions.

\textbf{Layers 18--27, successes:} Restoring the baseline residual stream \emph{increases} target probability, slightly at first and sharply at the last layers ($+0.24$ at layer 26, $+0.38$ at layer 27). In successes, the instruction's suppression is held in the late residual stream, and removing it undoes the suppression.

\textbf{Layers 20--27, failures:} Restoring the baseline residual stream \emph{decreases} target probability (down to $-0.15$ at layer 24). In failures, the late residual stream under the instruction holds \emph{more} of the target than under the baseline, consistent with the instruction having primed it.

Patching only the attention output (type A) has much smaller effects ($|\Delta P| \leq 0.10$), so the effect is not carried by any single layer's attention output at this position. This pattern provides direct causal evidence that layers 20--27 carry the difference between compliance and violation, in opposite directions for the two outcomes.

%==============================================================================
\section{A Taxonomy of Failure Modes}
\label{sec:taxonomy}
%==============================================================================

The mechanistic evidence suggests that not all failures are alike. We test two candidate modes with quantifiable signatures, and for each we measure how often the signature also appears in successes. A signature that is as common in successes as in failures does not explain failure.

\subsection{Priming Failure (27.2\%, $n = 170$)}

\textbf{Definition.} The instruction's explicit mention of $X$ activates rather than suppresses the target. The word ``Paris'' in ``do not say Paris'' acts as a cue.

\textbf{Signature.} $P_1 > P_0$: the instruction \emph{increased} the target's probability. We also test the attention signature $\text{TMF} > \text{NF}$ (more attention to the target mention than to the negation, averaged over layers), and the combination of the two.

\textbf{Base rates.} Table~\ref{tab:signatures} gives each signature's rate among failures and successes. The probability signature is twice as common in failures (27.2\% vs.\ 12.8\%; difference $+14.4$ points, 95\% CI $[10.9, 18.0]$). The attention signature is \emph{less} common in failures than in successes (89.6\% vs.\ 92.6\%). It therefore cannot be a mechanism of failure, and neither can the combination, which is dominated by it.

\subsection{Override Failure (39.9\%, $n = 249$)}

\textbf{Definition.} Suppression exists ($\Delta P > 0.1$) but is insufficient: the late-layer FFN push toward the target (Section~4.4) outweighs it.

\textbf{Signature.} Positive suppression magnitude ($\Delta P > 0.1$) together with a violating greedy answer. This holds for 39.9\% of failures; the remaining 32.9\% show little change ($0 \leq \Delta P \leq 0.1$). Among successes, 54.1\% show $\Delta P > 0.1$, as expected when suppression works.

\begin{table*}[t]
\centering
\caption{Rates of candidate failure signatures among failures ($n=624$) and successes ($n=1{,}876$). CI: prompt bootstrap; $p$: Fisher's exact test.}
\label{tab:signatures}
\begin{tabular}{@{}lrrrr@{}}
\toprule
Signature & Failures & Successes & Difference (95\% CI) & $p$ \\
\midrule
$P_1 > P_0$ & 27.2\% & 12.8\% & $+14.4$ $[10.9, 18.0]$ & $<10^{-15}$ \\
$\text{TMF} > \text{NF}$ & 89.6\% & 92.6\% & $-3.0$ $[-5.7, -0.5]$ & 0.02 \\
Either (priming rule) & 94.1\% & 94.4\% & $-0.3$ $[-2.5, 1.7]$ & 0.76 \\
\bottomrule
\end{tabular}
\end{table*}

\subsection{Implications of the Taxonomy}

The probability signature has a clear implication: \textbf{explicitly naming the forbidden word is a risk factor.} In just over a quarter of failures, naming the target raised its probability; at low pressure, where failures are rare, this is the typical failure. Not naming the target should reduce these failures.

Alternative formulations might include:
\begin{itemize}
    \item Category-level constraints: ``Do not mention any cities'' instead of ``Do not say Paris''
    \item Positive reframing: ``Use only general geographic terms''
    \item Circumlocution: ``Avoid the word that rhymes with `ferris'\,''
\end{itemize}

We leave systematic evaluation of alternative phrasings to future work, but our analysis provides a principled basis for expecting them to help with priming failures, though not with override failures.

%==============================================================================
\section{Discussion}
%==============================================================================

\subsection{Summary}

We have traced negative instruction failure from behavioral regularity to computational mechanism:

\begin{enumerate}
    \item \textbf{Behavioral level:} Semantic pressure (baseline probability) predicts the average violation rate via a logistic relationship (slope 2.76), though single prompts vary widely around it.
    
    \item \textbf{Suppression level:} At equal pressure, negative instructions suppress the target far less in failures than in successes, and at low pressure they raise it in failures.
    
    \item \textbf{Attention level:} Attention to the target mention versus the negation does not distinguish failures from successes.
    
    \item \textbf{Component level:} In late layers (22--27), attention suppresses while FFN promotes the target. In failures, the FFN push is larger and the attention push against the target smaller.
    
    \item \textbf{Causal level:} Activation patching shows the late-layer residual stream (layers 20--27) carries the instruction's effect, in opposite directions for successes and failures.
    
    \item \textbf{Taxonomic level:} Priming ($P_1 > P_0$) accounts for 27\% of failures and is twice as common as in successes; failures with real but insufficient suppression account for 40\%.
\end{enumerate}

This gives a mechanistic account of one instruction-following failure in one model, with each proposed signature tested against its base rate.

\subsection{Implications for Practitioners}

\textbf{Avoid explicit target naming.} In over a quarter of failures, mentioning the forbidden word raised its probability. Alternative phrasings that avoid naming the target directly may prove more effective.

\textbf{Anticipate difficulty for natural completions.} Constraints against the most natural answer are inherently hardest. Systems should estimate semantic pressure before deployment and apply additional safeguards for high-pressure cases.

\textbf{Post-generation filtering for critical applications.} Generation-time constraints alone may be insufficient. For safety-critical applications, combine generation constraints with output filtering.

\textbf{Monitor probabilities, not attention.} The attention-based Priming Index does not predict failure. Comparing the target's probability with and without the constraint ($P_1$ vs.\ $P_0$) does, and is cheap to compute.

\subsection{Limitations}

We acknowledge limitations that qualify our conclusions:

\textbf{Single model.} All experiments use Qwen2.5-7B-Instruct. Results may differ for:
\begin{itemize}
    \item Significantly larger models (70B+), which may have different layer dynamics
    \item Different architectures (Mixture of Experts, state-space models)
    \item Different training procedures (RLHF variants, Constitutional AI)
\end{itemize}

\textbf{Single-token focus.} We analyze the first token of multi-token targets. Multi-token dynamics may differ, especially for targets requiring sequential commitment. On the 1,071 prompts whose preferred spelling is a single token, the main mechanistic results hold: the pooled suppression gap is $-0.67$ (95\% CI $[-0.70, -0.64]$), and $P_1 > P_0$ occurs in 30\% of failures vs.\ 12\% of successes.

\textbf{Numerical precision.} The main run uses bfloat16. The same prompt evaluated in different batches gives next-token probabilities that differ by a median of 0.003 (90th percentile 0.04, maximum 0.18 over 250 prompts). This noise can move prompts near a bin edge or the success boundary; it is small relative to the effects reported.

\textbf{Controlled task setting.} Our experiments use one-word completions with unambiguous targets. Naturalistic settings (longer outputs, ambiguous constraints) may exhibit different patterns.

\textbf{Attention interpretability.} Attention weights are routing signals, not direct indicators of information use. The causal role of attention patterns requires targeted knockout experiments beyond our scope.

\textbf{Patching resolution.} We patch the residual stream and the attention output at one position. Finer-grained patching (individual heads, FFN neurons, other positions) could localize responsibility more precisely.

\textbf{Outcome-defined comparisons.} Failure is defined by the target appearing, which requires a high $P_1$, so part of the suppression gap between outcomes follows from the definition. Comparing within pressure bins removes the confound with $P_0$ but not this one.

\subsection{Future Directions}

\textbf{Cross-model validation:} Do pressure-violation relationships and failure modes generalize across architectures?

\textbf{Alternative phrasings:} Systematic comparison of constraint formulations to identify strategies that minimize priming.

\textbf{Head-level patching:} Which specific attention heads process negation versus target activation?

\textbf{Safety-relevant constraints:} Application to understanding failures of safety-oriented negative constraints.

%==============================================================================
\section*{Conclusion}
%==============================================================================

We have shown that negative instruction failure is neither random nor mysterious. It follows a computational tension: semantic pressure pulling toward a natural completion versus constraint pressure pushing away. When the former exceeds the latter---and sometimes because the constraint \emph{primes} the target it intends to suppress---failure follows.

This understanding opens paths toward more robust constraint design. Where naming the target primes it, avoid naming it. Where late layers fail to complete the suppression, target interventions there. The black box, opened, reveals not chaos but mechanism---and mechanism suggests remedy.

\section*{Reproducibility}

All code, data, and analysis scripts are publicly available.\footnote{\url{https://github.com/gut-puncture/Semantic-Gravity-RP}} The repository includes: the snapshotted sources, all judged candidates and the 2,500 selected prompts with targets, categories and pressure; the complete measurement and analysis pipeline; the per-prompt outputs; and the scripts that generate every figure and table number from them. Golden checks run with every experiment: the logit lens at the last layer matches the model's output, $P_0$ matches a brute-force estimate, patching nothing is exact, token spans decode to the intended text, and the attribution terms sum to the final logit. Key parameters: temperature 1.0, top-$p$ 0.9, 16 samples/prompt, seed 42. Detection is deterministic.

\textbf{Changes from version 1.} Version 1 of this paper contained errors that this version corrects: the pressure measure summed overlapping variants and was clipped to 1; the logit lens applied the final normalization twice at the last layer; activation patching reused a cached state that earlier patches had modified; confidence intervals resampled samples rather than prompts; prompts were given without the chat template and batched sampling did not use left padding; the failure taxonomy had no base rate among successes; and the stated pressure filter ($P_0 \geq 0.2$) was not applied. All experiments were rerun with corrected code, and every number in this version is computed from the saved outputs.

\newpage
%==============================================================================
\appendix
\section*{Appendix}
%==============================================================================

\section{Dataset Construction Details}
\label{app:dataset}

Every external source is saved with its retrieval date, and every prompt records its source. All randomness uses seed 42. A target word appears in at most two categories.

\textbf{Idiom prompts.} Idioms from the \texttt{english\_idioms} dataset (baiango, pinned commit), supplemented by Wiktionary idioms where the first source ran short, with the final word blanked using fixed templates. Examples: ``A bird in the hand is worth two in the \_\_\_'' (bush); ``Don't count your chickens before they \_\_\_'' (hatch).

\textbf{Factual prompts.} Wikidata SPARQL queries for relations with unique values (for example capitals, currencies and chemical elements), saved as JSON and converted to questions with fixed templates. Examples: ``The chemical symbol for gold is \_\_\_'' (Au); ``The largest planet in our solar system is \_\_\_'' (Jupiter).

\textbf{Common sense prompts.} We first built questions from ConceptNet assertions, but few passed the blind judge (46 accepted). We therefore generated common-sense questions with an LLM (GPT-6.1-terra), one request per target word asking for five candidates, and judged them with the same blind rules. Examples: ``You use scissors to \_\_\_ paper'' (cut); ``A refrigerator keeps food \_\_\_'' (cold).

\textbf{Creative/OOD prompts.} Target words are chosen first (word frequency between Zipf 3.5 and 6.0, alphabetic, distinct across the generated categories). One LLM request per word (GPT-6.1-terra) asks for five candidate prompts, so duplicates cannot arise. Among the accepted candidates for a word, one is kept at random, not the one with the highest $P_0$: choosing by $P_0$ would select prompts on the quantity being studied.

\textbf{Multi-token targets.} Targets may span several tokens (many capitals do). The mechanistic measures use the target's first token, and each mechanistic result is also reported on the subset of prompts whose preferred spelling is a single token.

\section{Semantic Pressure Computation}
\label{app:pressure}

For each target $X$, we generate variants:
\begin{itemize}
    \item Capitalization: as given, lower, Title, UPPER (duplicates removed)
    \item Whitespace: with and without a leading space (per BPE tokenizer behavior)
\end{itemize}

Each variant is tokenized canonically, and its probability is computed by a teacher-forced forward pass. It is then multiplied by the probability that the next token does not start with a letter or digit, so punctuation after the word is allowed but a longer word is not. Because each variant is a distinct complete first word, the sum over variants is a probability and never exceeds 1; we check this for every prompt. A golden check compares $P_0$ with a brute-force estimate from sampled completions.

\section{Extended Results Tables}

\begin{table}[H]
\centering
\caption{Violation rate by pressure bin ($n=40{,}000$ samples; $n$ = prompts; CI: prompt bootstrap).}
\begin{tabular}{@{}lrrrr@{}}
\toprule
Pressure Bin & Mean & CI Low & CI High & $n$ \\
\midrule
$[0.0, 0.2)$ & 0.054 & 0.044 & 0.066 & 1,004 \\
$[0.2, 0.4)$ & 0.141 & 0.098 & 0.183 & 165 \\
$[0.4, 0.6)$ & 0.214 & 0.165 & 0.266 & 146 \\
$[0.6, 0.8)$ & 0.219 & 0.183 & 0.259 & 282 \\
$[0.8, 1.0]$ & 0.439 & 0.413 & 0.467 & 903 \\
\bottomrule
\end{tabular}
\end{table}

\begin{table}[H]
\centering
\caption{FFN and attention contributions at layer 27 (direct logit attribution, logits; negative instruction condition, $P_0 \geq 0.8$).}
\begin{tabular}{@{}lrr@{}}
\toprule
Outcome & Attn Contrib & FFN Contrib \\
\midrule
Success ($n=465$) & $-1.18$ & $+9.66$ \\
Failure ($n=438$) & $-0.78$ & $+12.43$ \\
\bottomrule
\end{tabular}
\end{table}

\begin{table}[H]
\centering
\caption{Residual-stream patching effects by layer (type R; 250 prompts, 50 per pressure bin).}
\begin{tabular}{@{}lrr@{}}
\toprule
Layers & Successes & Failures \\
\midrule
0--17 & $|\Delta P| < 0.01$ & $|\Delta P| < 0.01$ \\
18--19 & $+0.01$ to $+0.03$ & $-0.01$ to $0.00$ \\
20--23 & $+0.01$ to $+0.06$ & $-0.05$ to $-0.14$ \\
24--25 & $+0.02$ to $+0.07$ & $-0.14$ to $-0.15$ \\
26--27 & $+0.24$ to $+0.38$ & $-0.11$ to $-0.14$ \\
\bottomrule
\end{tabular}
\end{table}

\vfill

% References - NeurIPS allows 9pt font for references
{\small
\bibliographystyle{plainnat}
\bibliography{references}

@inproceedings{ouyang2022training,
  title={Training Language Models to Follow Instructions with Human Feedback},
  author={Ouyang, Long and Wu, Jeffrey and Jiang, Xu and Almeida, Diogo and Wainwright, Carroll and Mishkin, Pamela and Zhang, Chong and Agarwal, Sandhini and Slama, Katarina and Ray, Alex and Schulman, John and Hilton, Jacob and Kelton, Fraser and Miller, Luke and Simens, Maddie and Askell, Amanda and Welinder, Peter and Christiano, Paul and Leike, Jan and Lowe, Ryan},
  booktitle={Advances in Neural Information Processing Systems},
  volume={35},
  pages={27730--27744},
  year={2022},
  publisher={Curran Associates, Inc.}
}

@inproceedings{wei2022finetuned,
  title={Finetuned Language Models are Zero-Shot Learners},
  author={Wei, Jason and Bosma, Maarten and Zhao, Vincent and Guu, Kelvin and Yu, Adams Wei and Lester, Brian and Du, Nan and Dai, Andrew M and Le, Quoc V},
  booktitle={International Conference on Learning Representations},
  year={2022},
  url={https://openreview.net/forum?id=gEZrGCozdqR}
}

@inproceedings{kassner2020negated,
  title={Negated and Misprimed Probes for Pretrained Language Models: Birds Can Talk, But Cannot Fly},
  author={Kassner, Nora and Sch{\"u}tze, Hinrich},
  booktitle={Proceedings of the 58th Annual Meeting of the Association for Computational Linguistics},
  pages={7811--7818},
  year={2020},
  publisher={Association for Computational Linguistics}
}

@inproceedings{hossain2022analysis,
  title={An Analysis of Negation in Natural Language Inference},
  author={Hossain, Md Mosharaf and Blanco, Eduardo and Kautz, Henry},
  booktitle={Proceedings of the 2022 Conference on Empirical Methods in Natural Language Processing},
  pages={5383--5395},
  year={2022},
  publisher={Association for Computational Linguistics}
}

@article{nostalgebraist2020logitlens,
  title={Interpreting {GPT}: The Logit Lens},
  author={nostalgebraist},
  journal={LessWrong},
  year={2020},
  note={Available at \url{https://www.lesswrong.com/posts/AcKRB8wDpdaN6v6ru/interpreting-gpt-the-logit-lens}}
}

@inproceedings{geva2022transformer,
  title={Transformer Feed-Forward Layers Build Predictions by Promoting Concepts in the Vocabulary Space},
  author={Geva, Mor and Caciularu, Avi and Wang, Kevin and Goldberg, Yoav},
  booktitle={Proceedings of the 2022 Conference on Empirical Methods in Natural Language Processing},
  pages={30--45},
  year={2022},
  publisher={Association for Computational Linguistics}
}

@inproceedings{meng2022locating,
  title={Locating and Editing Factual Associations in {GPT}},
  author={Meng, Kevin and Bau, David and Andonian, Alex and Belinkov, Yonatan},
  booktitle={Advances in Neural Information Processing Systems},
  volume={35},
  pages={17359--17372},
  year={2022},
  publisher={Curran Associates, Inc.}
}

@inproceedings{wang2023interpretability,
  title={Interpretability in the Wild: A Circuit for Indirect Object Identification in {GPT-2} Small},
  author={Wang, Kevin and Variengien, Alexandre and Conmy, Arthur and Shlegeris, Buck and Steinhardt, Jacob},
  booktitle={International Conference on Learning Representations},
  year={2023},
  url={https://openreview.net/forum?id=NpsVSN6o4ul}
}

@inproceedings{zhao2021calibrate,
  title={Calibrate Before Use: Improving Few-Shot Performance of Language Models},
  author={Zhao, Zihao and Wallace, Eric and Feng, Shi and Klein, Dan and Singh, Sameer},
  booktitle={Proceedings of the 38th International Conference on Machine Learning},
  pages={12697--12706},
  year={2021},
  publisher={PMLR}
}

@inproceedings{lu2022fantastically,
  title={Fantastically Ordered Prompts and Where to Find Them: Overcoming Few-Shot Prompt Order Sensitivity},
  author={Lu, Yao and Bartolo, Max and Moore, Alastair and Riedel, Sebastian and Stenetorp, Pontus},
  booktitle={Proceedings of the 60th Annual Meeting of the Association for Computational Linguistics},
  pages={8086--8098},
  year={2022},
  publisher={Association for Computational Linguistics}
}

@article{elhage2021mathematical,
  title={A Mathematical Framework for Transformer Circuits},
  author={Elhage, Nelson and Nanda, Neel and Olsson, Catherine and Henighan, Tom and Joseph, Nicholas and Mann, Ben and Askell, Amanda and Bai, Yuntao and Chen, Anna and Conerly, Tom and DasSarma, Nova and Drain, Dawn and Ganguli, Deep and Hatfield-Dodds, Zac and Hernandez, Danny and Jones, Andy and Kernion, Jackson and Lovitt, Liane and Ndousse, Kamal and Amodei, Dario and Brown, Tom and Clark, Jack and Kaplan, Jared and McCandlish, Sam and Olah, Chris},
  journal={Transformer Circuits Thread},
  year={2021},
  url={https://transformer-circuits.pub/2021/framework/index.html}
}
}

\end{document}